\documentclass[11pt,a4paper]{article}
\usepackage{times,latexsym}
\usepackage{url}
\usepackage[T1]{fontenc}
\usepackage{xcolor}  

\usepackage[acceptedwithA]{tacl2021v1}

\usepackage{xspace,mfirstuc,tabulary}

\usepackage[utf8]{inputenc} 
\usepackage{microtype} 
\usepackage{inconsolata} 
\usepackage{graphicx} 
\usepackage{listings} 
\usepackage{lipsum} 
\usepackage{mwe} 
\usepackage{todonotes} 
\usepackage{amsmath} 
\usepackage{enumitem}
\usepackage{booktabs}
\usepackage{multirow}
\usepackage{float}
\usepackage{tcolorbox}

\newcommand{\mrcad}{mrCAD}
\newcommand{\designer}{\textit{Designer}}
\newcommand{\Designer}{\textit{Designer}}
\newcommand{\maker}{\textit{Maker}}
\newcommand{\Maker}{\textit{Maker}}

\lstdefinestyle{mrcadlststyle}{
    basicstyle=\ttfamily\footnotesize, 
    breakatwhitespace=false,         
    breaklines=true,                 
    captionpos=b,                    
    keepspaces=true,                 
    numbersep=5pt,                  
    showspaces=false,                
    showstringspaces=false,
    showtabs=false,                  
    tabsize=2
}

\lstdefinestyle{pythonstyle}{
    language=Python,
    basicstyle=\ttfamily\small,
    keywordstyle=\color{blue}\bfseries,
    commentstyle=\color{gray},
    stringstyle=\color{red},
    numbers=left,
    numberstyle=\tiny,
    stepnumber=1,
    breaklines=true,
    breakatwhitespace=true,
    showstringspaces=false,
    frame=tb, 
}

\newif\iftaclinstructions
\taclinstructionsfalse 
\iftaclinstructions
\renewcommand{\confidential}{}
\renewcommand{\anonsubtext}{(No author info supplied here, for consistency with
TACL-submission anonymization requirements)}
\newcommand{\instr}
\fi

\iftaclpubformat 

\else

\fi

\title{\raisebox{-0.5em}{\includegraphics[height=1.5em]{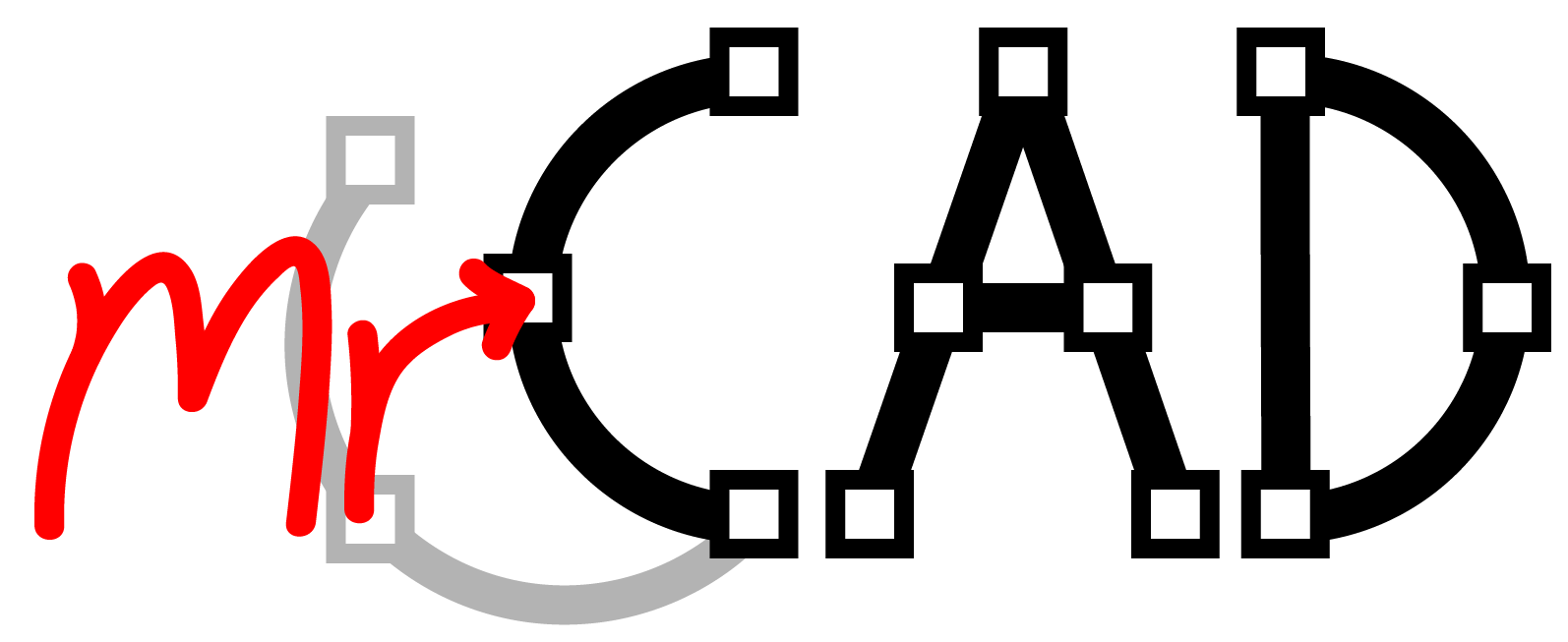}} Multimodal Refinement of Computer-aided Designs}

\author{William P. McCarthy$^1$\Thanks{equal contribution}\;\;\;Saujas Vaduguru$^{2*}$\Thanks{work partly done as an intern at Autodesk AI Lab}\;\;\;Karl D. D. Willis$^1$\;\;\;Justin Matejka$^1$ \\ \textbf{Judith E. Fan}$^3$\;\;\;\textbf{Daniel Fried}$^2$\;\;\;\textbf{Yewen Pu}$^1$ \\
$^1$Autodesk AI Lab\;\;\;$^2$Carnegie Mellon University\;\;\;$^3$Stanford University}

\date{}

\begin{document}
\maketitle
\begin{abstract}
A key feature of human collaboration is the ability to iteratively refine the concepts we have communicated.
In contrast, while generative AI excels at the \textit{generation} of content, it often struggles to make specific language-guided \textit{modifications} of its prior outputs.
To bridge the gap between how humans and machines perform edits, we present \mrcad{}, a dataset of multimodal instructions in a communication game\footnote{dataset repo: \url{https://github.com/AutodeskAILab/mrCAD}}. 
In each game, players created computer aided designs (CADs) and refined them over several rounds to match specific target designs.
Only one player, the \designer{}, could see the target, and they must instruct the other player, the \maker{}, using text, drawing, or a combination of modalities.
\mrcad{} consists of 6,082 communication games, 15,163 instruction-execution rounds, played between 1,092 pairs of human players.
We analyze the dataset and find that generation and refinement instructions differ in their composition of drawing and text. 
Using the \mrcad{} task as a benchmark, we find that state-of-the-art VLMs are better at following generation instructions than refinement instructions.
These results lay a foundation for analyzing and modeling a multimodal language of refinement that is not represented in previous datasets.
\end{abstract}

\section{Introduction}

Recent advances in generative AI have enabled collaborative design via a natural language interface, with applications such as generating images and code.
However, while these models demonstrate impressive results in generating high-quality content, they struggle to \textit{modify} what they have generated based on further user refinement instructions. 
The challenge of getting models to perform interactive edits is well-documented in tasks such as image editing \cite{fu2024guiding} and code generation \cite{cassano2023can}.
This often results in frustration with AI-generated content that is ``nearly there'' but difficult to refine to exactly how the user wanted it.

\begin{figure}[t]
  \centering
  \includegraphics[width=0.45\textwidth]{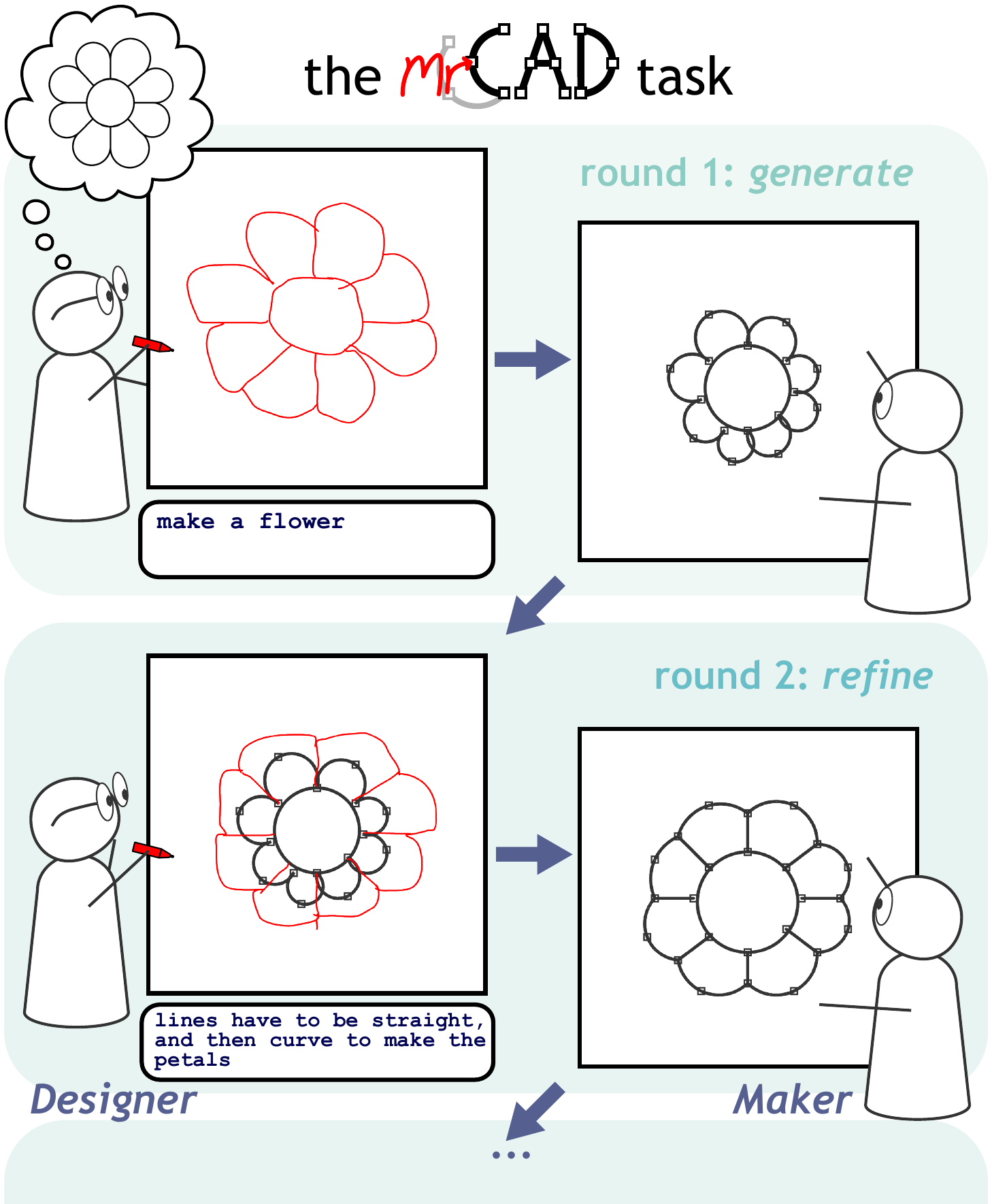} 
  \caption{We present a dataset of multimodal instructions for creating and modifying computer aided designs, along accompanying executions of these instructions in a 2D CAD environment. Participants worked together to recreate a target CAD over multiple rounds. Crucially, the target design is known only to the designer, who must instruct the maker on what to create. The maker, in turn, must edit the CAD model based on these instructions.}
  \vspace{-1em}
  \label{fig:opening}
\end{figure}

On the other hand, humans regularly instruct each other to perform refinements.
Works such as \citet{dingemanse2024interactive} have argued that since human communication is infinitely expressive, repair is a core mechanism for ensuring robust communication.
Further, human-given refinements are often multimodal \cite{lawson2006designers, williams2002designers} where different modalities interact to make the instruction precise.

We present \mrcad{}, a resource for studying and benchmarking multi-modal refinement in a computer-aided design (CAD) setting. 
An example data-point is shown in Figure \ref{fig:opening}. 
One player, the \emph{Designer}, has a target design which they demonstrate via a combination of text and drawing instructions. The other player, the \emph{Maker}, uses a programmatic CAD language to carry out the Designer's instructions and attempt to recreate the target. In subsequent rounds, the Designer and Maker iteratively \emph{refine} the design to get closer to the target.

Our task setup produces both (1) natural interaction between modalities: e.g.\ in Figure 1, the text specifies `lines have to be straight' and the drawing specifies \emph{which} lines need to be straight and (2) contextual dependence, where refinement instructions need to be interpreted in the context of what has been constructed so far.
People naturally use both types of context when generating and interpreting instructions (Sec.\ \ref{sec:effects-of-multimodality}); but we find that models struggle particularly with refinement (Sec.\ \ref{sec:evaluation}).

\paragraph{Benchmarking} As a grounded instruction-following benchmark, \mrcad{} benefits from the programmatic representation of CAD.
CAD reconstruction accuracy can be programmatically defined, in contrast to image creation tasks, which rely on learned metrics such as CLIP~\cite{radford2021learning}.
Additionally, CAD programs can be easily manipulated using a sequence of simple clicks and drags, making it easy to evaluate both tool-using agents and human participants in an evaluation environment.

\paragraph{Understanding} As a resource for understanding the nature of multimodal refinement in human-human collaboration, 
\mrcad{} is large scale, containing 6,082 human-to-human plays of the game, with a total of 15,163 instruction-execution rounds. 
A large number of CADs in our dataset are recreated \emph{multiple} times by different dyads, ranging from 2 to 30 games per CAD design. 
This way, we can understand the variation in refinement strategies across people. 
The CAD designs in \mrcad{} are naturalistic, sourced from the SketchGraphs dataset \cite{seff2020sketchgraphs}, consisting of 2D CAD created by designers.
A notable feature of CAD is that while it only contains simple curves of lines, circles, and arcs, put together these simple curves form rich semantic objects and sub-parts (like the ``flower'' and ``petals'' in Figure \ref{fig:opening}), which must be parsed contextually \cite{ji2022abstract}.
This is in contrast to other scene creation datasets such as \citet{zitnick2013bringing}, where each object in the scene has a canonical, pre-defined name (e.g. a ``boy'' and a ``ball'').
The instructions we collect in \mrcad{} consist of freeform drawings and text, highlighting how humans leverage multimodality and refinement in an open-ended manner.

Our work makes the following contributions:
\begin{itemize}[]
    \item A novel task of multi-modal refinement of CAD designs.
    \item A detailed procedure for collecting a dataset of multi-modal refinement communications using crowd-sourcing.
    \item The dataset and benchmark, consisting of 15,163 instructions and executions, wrapped in an accessible gym environment.
    \item Analyses of human-human communication that reveal differences in the multimodal languages of generation and refinement.
    \item Evaluation of existing VLM models, revealing a severe gap in their refinement ability, particularly compared to their generation ability.
\end{itemize}

\section{The \mrcad{} Task}

The \mrcad{} task (Fig. \ref{fig:opening}) is a two player, multi-turn communication game.
Following \citet{mccarthy2024communicating}, a \designer{} issues multimodal instructions to guide a \maker{} to take actions in a mutually observable CAD environment to construct a target CAD.
The target CAD is known only to the \designer{}.

\subsection{\mrcad{} environment} \label{sec:env}

\paragraph{State} The state $D$ is a CAD data structure.
\begin{lstlisting}
D -> {Curve, ...}
Curve  -> Line | Circle | Arc 
Line   -> l(P,P) // end points
Circle -> c(P,P) // points on diameter
Arc    -> a(P,P,P) // start, mid, end
\end{lstlisting}
Each curve is defined by its \emph{control points} that the \maker{} selects. 
A \texttt{render} function allows players to view the state by rendering the design as an image.

\paragraph{Action} The \maker{} can alter the CAD by making, removing, and moving curves via translation.
They can also change the shape of a curve by moving its control points. 
If multiple curves share the same control point, moving the point would modify all the curves, and deleting the point would delete all the curves.
\begin{lstlisting}[mathescape=true]
Action -> make_curve Curve 
        | remove_curve Curve 
        | move_curve Curve Vxy
        | move_point P P
        | delete_point P
\end{lstlisting}


\paragraph{Transition} A sequence of actions $A = [a_1, a_2, \ldots, a_n]$ can be applied to a design, which results in a new design.
\begin{align*}
    D' &= A(D) = [a_1, a_2, \ldots, a_n](D) \\
    &= (a_n \circ \cdots \circ a_1)(D)
\end{align*}

\begin{figure}[t]
  \centering
  \includegraphics[width=0.3\textwidth]{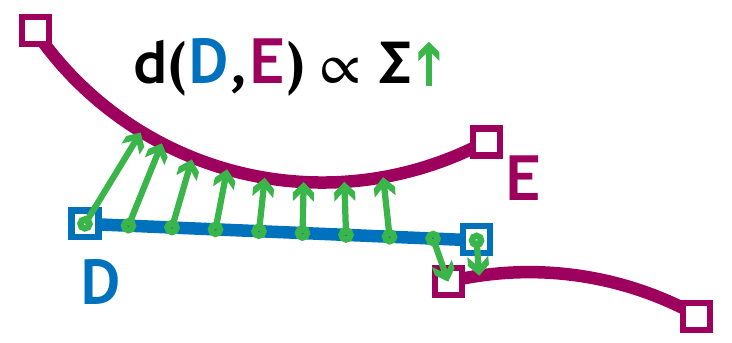} 
  \caption{The \textbf{asymmetric chamfer distance} from CAD $D$ to CAD $E$ is calculated by sampling 10 points on every curve of $D$, and calculating (symbolically) the minimum distance from each point to $E$.
  Each distance is then normalized by multiplying $\frac{1}{4}$ of the maximum size of the canvas, making it invariant to the canvas size.
  These distances are summed, making the asymmetrical chamfer distance.
  The symmetric Chamfer distance we use is the average of both directions.
}
\vspace{-1em}
  \label{fig:dist_metric}
\end{figure}

\paragraph{Distance Metric} 
CADs are programmatic entities that exist in a visual space, where very different elements could be used to create two CADs of similar appearance.
For example, a series of straight lines can be used to represent a curve.
Therefore, we devise a distance metric $\Delta$ based on chamfer distance \cite{butt1998optimum} that respects the vector-like nature of CADs, while accounting for geometric similarity (Fig. \ref{fig:dist_metric}).


\paragraph{Instructions}
The \designer{} can instruct the \maker{} using multimodal messages.
\begin{lstlisting}[mathescape=true]
Message -> $\langle$Text, Drawing$\rangle$
Text    -> [char, ...] | empty
Drawing -> [stroke, ...] | empty
\end{lstlisting}
A text is a sequence of characters and a drawing is a sequence of strokes, encoded in SVG format.

\subsection{Playing the Game}
The shared goal of the \designer{} and \maker{} is to collaboratively reconstruct a target design, which is only given to the Designer.

\paragraph{Round and Rollout} A round is a tuple of: a design $D$, the message from the designer $m$, the actions generated from the maker $A$, and the resulting updated design $D' = A(D)$. A rollout is a sequence of rounds.
\begin{align*}
r_i &= (D_i, m_i, A_i, D'_i) \\
R &= [r_1, \ldots, r_n]
\end{align*}    

Each rollout has $D_1 = \{\}$ and $D_i = D'_{i - 1}$.
A game is ``won'' if the final design is within a certain threshold $\theta$ of distance from the target $D^*$.
    \begin{align*}
        \Delta(D'_n, D^*) < \theta
    \end{align*}

\paragraph{\Designer} At round $i$, the designer is given both the target and current designs, rendered as images f $D^*$ and $D_i$, along with the history of the (rendered) interaction of previous rounds $\texttt{render}(R_{1:i-1})$, and generates a message $m$ from 
    \begin{align*}
        P_\text{designer}(m &~|~\texttt{render}(D^*), \\
        & \texttt{render}(D_i), \texttt{render}(R_{1:i-1}))
    \end{align*}

Presenting designer only the rendered CADs encourages them to focus on the geometric properties and communicate the refinements naturalistically, instead of communicating about the underlying CAD program itself.

\paragraph{\maker{}} The maker takes in the message, the current design, the interaction history, and generates a sequence of actions $A$ from
    \begin{align*}
        P_\text{maker}(A ~|~ m, D_i, R_{1:i-1})
    \end{align*}

\paragraph{Play}
Playing the game starts with  a target design $D^*$ and involves a dyad (a designer-maker pair) acting in turns to generate a rollout.
    \begin{align*}
        R = \texttt{play}(D^*, P_\text{designer}, P_\text{maker})
    \end{align*}
\section{Effects of Multimodality and Refinement on Communication}
\label{sec:effects-of-multimodality}

\begin{figure}[t]
  \centering
  \includegraphics[width=0.5\textwidth]{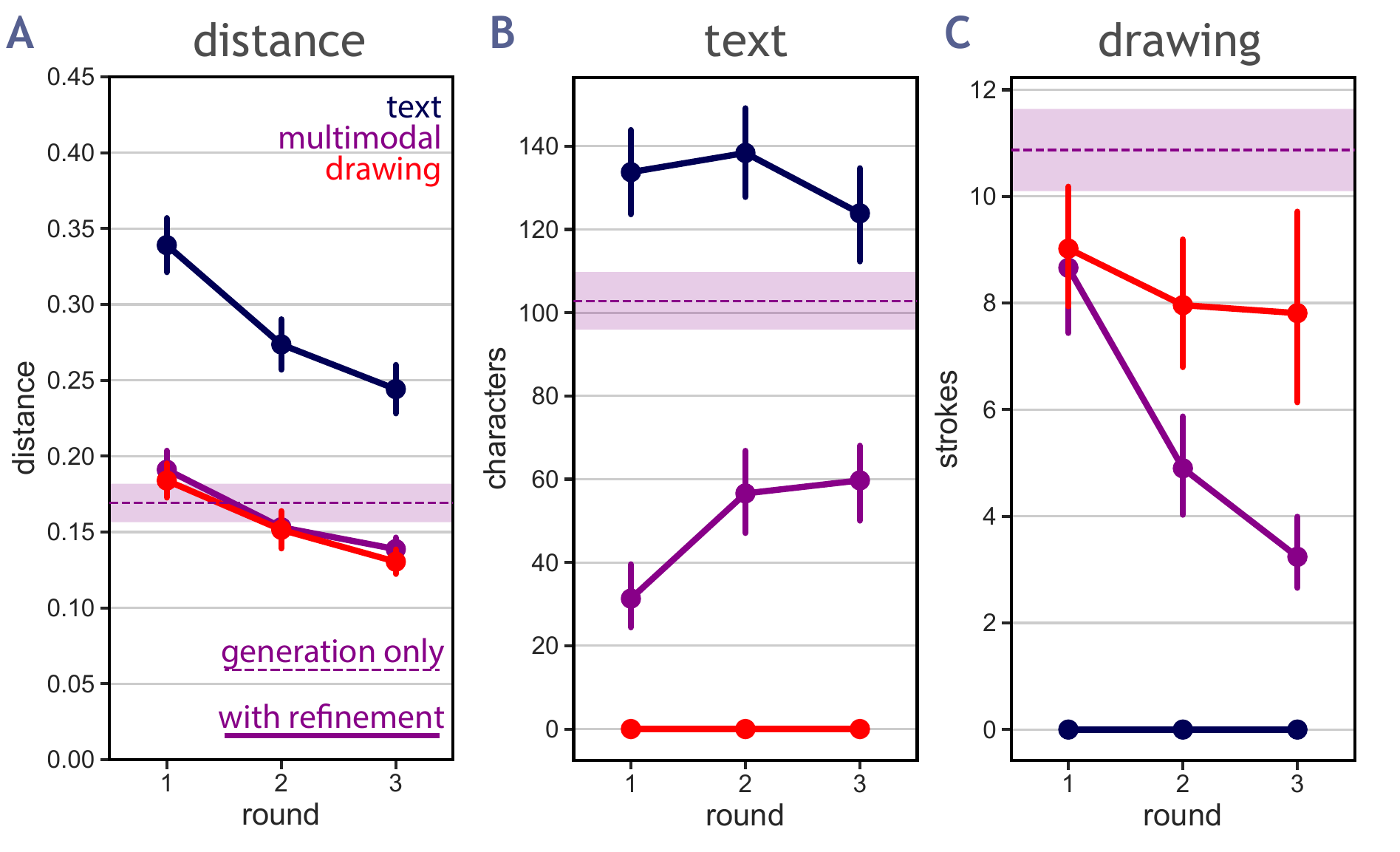} 
  \vspace{-2em}
  \caption{\textbf{A}: reconstruction accuracy for the 4 communication conditions --- multimodal+refinement, text only + refinement, drawing only + refinement, and multimodal + generation only. Using text only was less effective. Having no refinement was less effective.
  With refinement, drawing only and multimodal are comparable in performance.
  \textbf{B}: usage of text across rounds --- in the multimodal condition, participants used more texts in the later refinement rounds, suggesting a usage of text in conjunction with drawings to communicate refinements.
  \textbf{C}: usage of drawing across rounds --- in the multi-modal condition, participants used more drawing in the generation round, and less in the refinement rounds.
  }
  \label{fig:experiments}
\end{figure}

How do multimodality and refinement each support efficient communication?
Prior to collecting our large-scale dataset, we conducted a set of human ablation studies to answer these questions, manipulating the availability of (1) communication modalities, and (2) refinement rounds.


\subsection{Experimental methods}
Crowd source participants were paired and randomly assigned the role of \designer{} or \maker{}. 
We sampled 100 CADs from the Sketchgraph dataset \cite{seff2020sketchgraphs}, and presented these CADs across the set of dyads in each of four conditions:

\paragraph{\textit{multimodal + refinement}}
\designer{}'s messages could include text (up to 200 characters) and/or drawing instruction (unlimited). There are 3 rounds total, each round with 30 seconds for the \designer{} to construct a message, and 120 seconds for the \maker{} to manipulate the current CAD in a CAD interface.

\paragraph{\textit{text only + refinement}}
Where the message could only contain text but not drawing.

\paragraph{\textit{drawing only + refinement}}
Where the message could only contain drawing but not text.

\paragraph{\textit{multimodal + generation only}}
Here, rather than having 3 rounds of communication, they are aggregated into a single big round, with 90 seconds for the \designer{} and 360 seconds for the \maker{}.


\subsection{Results}

\paragraph{Refinement improves reconstruction}
Participants in all \textbf{refinement} conditions were able to improve their reconstructions over all three rounds ($b=-0.0335$, $t=-6.91$, $p=<0.001$) (Fig. \ref{fig:experiments}A).
While participants in the multimodal refinement condition on round one generated less accurate CADs than the \textbf{generation only} condition ($t=1.47$, $p = 0.145$), by round three their designs were reliably more accurate ($t=-2.15$, $p = 0.0340$), despite having the same total time available.
This highlights the value of iterative refinement for creating accurate designs.

\paragraph{Drawings improves reconstruction}
Participants who could only use text achieved considerably less accurate reconstructions than those in the drawing-only and multimodal conditions ($b=0.127$, $t=6.54$, $p=<0.001$).
While text is a relatively poor communicative medium on its own, participants in the multimodal condition still opted to use text in conjunction with drawings, and increased their use of text in later rounds (Fig. \ref{fig:experiments}B), suggesting that language is useful for \textit{refining} CADs.
The amount of drawing remained consistent in the drawing-only condition, but dropped precipitously after round 1 in the multimodal condition (Fig. \ref{fig:experiments}C).
Overall, when given a choice of multi-modality, human participants favored more drawing for the generation round, and more texts in the refinement rounds.

\paragraph{Interim discussion}
Why do participants choose to use multi-modal instructions when given a choice? 
Manually inspecting the data revealed that drawings are primarily used in two ways: Those used for creating a shape, and those used for editing a shape.
In the multimodal condition, the \designer{} can disambiguate these with texts such as ``make this shape'' or ``move the shape as shown''.
However, in the drawing only condition, the \designer{} frequently resorted to redrawing the entire target CAD from scratch --- a valid yet inefficient strategy.
Data collected from this experiment hint towards a complex interplay between linguistic and graphic communication that differs in generation and refinement settings.
The mrCAD dataset, discussed next, aims to provide a large scale dataset of \textit{multimodal generation and refinement instructions} to study these phenomena.


\section{Data Collection Procedure}

How can we scale-up the previous study to maximize the number of high-quality multimodal refinement instructions?
Ensuring quality in a dataset of this size involved multiple competing goals: collecting large volumes of high-quality data, minimizing costs, and fairly compensating crowd workers with varying degrees of effort and skill.


\subsection{mrCAD annotation task}
We recruited fluent English speakers from the USA and UK on Prolific.
Participants were paired, randomly assigned the role of \designer{} or \maker{}, and worked together over a series of rounds to recreate target CADs.
Rather than fixing a number of rounds and the amount of time per round, we instead limited the maximum round number to 10, and time to 9 minutes.
Within these limits, participants can choose how many rounds they will take, and to allocate different amount of times for each round.
We also lifted the limit on the amount of text characters the participants can send in a message.


\subsection{Sampling CAD tasks}
We sample tasks from the SketchGraphs dataset \cite{seff2020sketchgraphs}. 
We first normalize all designs to a 20 $\times$ 20 grid to identify duplicates. We then rescale the designs by a random scaling factor to ensure a variety of design sizes, while ensuring a minimum gap between pairs of elements such as control points, parallel lines, and concentric arcs or circles to ensure that they were far enough apart to be reproducible in the CAD construction interface. 
We then grouped designs into buckets based on a `signature' determined by the count of various types of curves in the design (horizontal lines, vertical lines, skewed lines, arcs, and circles). 
By selecting one design per bucket, we: (1) eliminated most near-duplicates (2) select a more diverse set of designs.
Participants were given 12 tasks randomly sampled from the task pool and shown in order of increasing difficulty (increasing number of curves in the design). 

\subsection{Prioritizing motivated annotators}
The \mrcad{} task is challenging.
Successful performance requires a person to (1) learn to operate a new CAD interface and (2) be adept at communicating CAD modifications, all within the span of one online session.
Furthermore, in a two player game, \emph{both} players must be motivated and capable for a successful reconstruction.
How can we prioritize capable dyads to provide high-quality annotations, while fairly compensating capable individuals with an unmotivated partner? 
We devised an incentive scheme as follows.

\paragraph{Prioritizing capable individuals}
Before participants started performing CAD tasks, they had to pass through three checks: a comprehension quiz, a solo CAD reconstruction task, and a paired practice trial.
In the solo CAD reconstruction task, each participant has to recreate a simple CAD consisting of a line, circle, and arc, to our fixed threshold level of accuracy.
All participants who passed the solo reconstruction task are deemed capable and motivated, and received a base pay of (\$8.70).

\paragraph{Prioritizing capable dyads}
Once paired, dyads had to successfully recreate one practice trial (a simple smiley face) in order to attempt other CADs.
We awarded a \$1 bonus per person for every recreation that met the accuracy threshold.
If a dyad fails to meet the threshold in 9 minutes, they lose a ``life''.
Dyads were ejected from the study when they lose 3 lives, limiting the number of CADs that unmotivated dyads could attempt.
This scheme directs our compensation towards high quality reconstructions from capable dyads.

\subsection{Prioritizing refinement instructions}
Our primary goal was to collect a large number of \textit{multimodal refinement instructions} that led to high-quality reconstructions.
A simple way of ensuring high-quality reconstructions it to implement a fixed threshold of accuracy for submitting CADs.
However, this fixed threshold does not take into account that some CAD would be easier to reconstruct while others would be more challenging.
In a pilot study, we found that with a lenient threshold, unmotivated dyads submitted low quality reconstructions \emph{without refinements}, while with a stringent threshold, motivated dyads were not compensated for decent quality reconstructions on more complex CADs.

Therefore, we implemented a \textbf{dynamic submission threshold}, which allowed only highly accurate reconstructions to be submitted in early rounds, but was more permissive in later rounds.
Implementing this change allowed us to collect 1,532 additional refinement rounds, an increase of 10.1\% compared to using a static threshold.
Furthermore, a dynamic threshold kept motivated dyads who had been ``stumped'' by a difficult CAD in the study longer by not losing a life.
The effect of dynamic threshold can be seen in Figure \ref{fig:splits}B.

\subsection{Exclusion criteria}
\begin{figure}[t]
  \centering
  \includegraphics[width=0.45\textwidth]{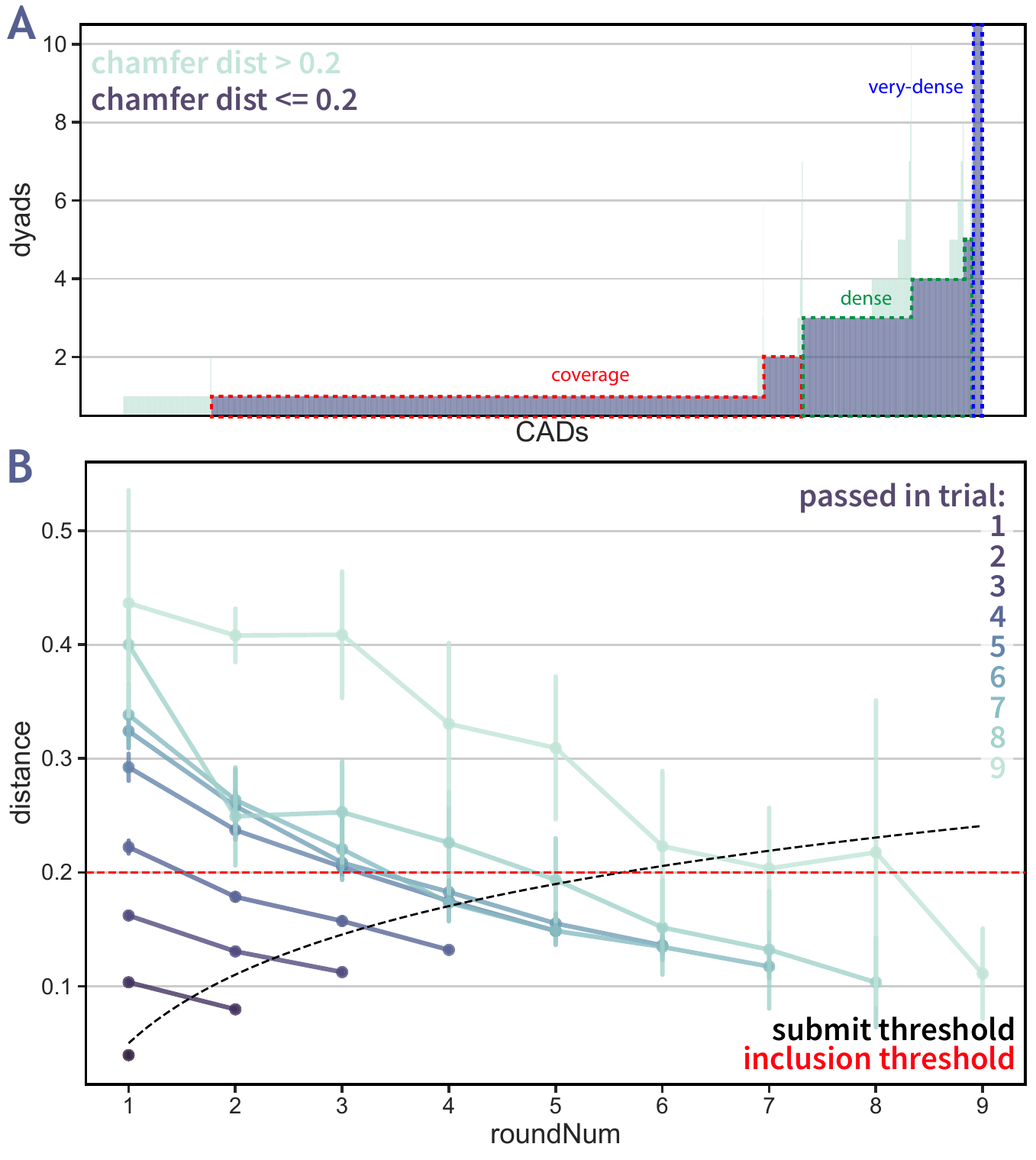} 
  \caption{
  \textbf{A} The \mrcad{} dataset contains three subsets: the \textbf{coverage set} of 2249 CADs with 1-2 successful rollouts, \textbf{dense set} of 698 CADs with 3+ successful reconstruction, and the \textbf{very-dense set} of 27 CADs with 30+ successful reconstruction.
  \textbf{B} We implemented a dynamic threshold for submitting designs that became more lenient in later rounds. Participants took a variable number of rounds to reach the threshold. Visualizing distance to the target broken down by round submitted reveals a trend of refinement over time.
  Red dashed line indicates the fixed threshold for including in analysis.
  }
  \label{fig:splits}
\end{figure}

Rollouts that contained no CAD actions, entirely empty messages, or had missing rounds were excluded from the dataset for analysis.
We also imposed a fixed \textbf{inclusion threshold} (Figure \ref{fig:splits}A) on reconstruction accuracy for analysis.
In the statistics below, we report only data that meets these criteria, however our full dataset release also includes failures and practice trials.



\begin{figure*}
  \centering
  \includegraphics[width=1\textwidth]{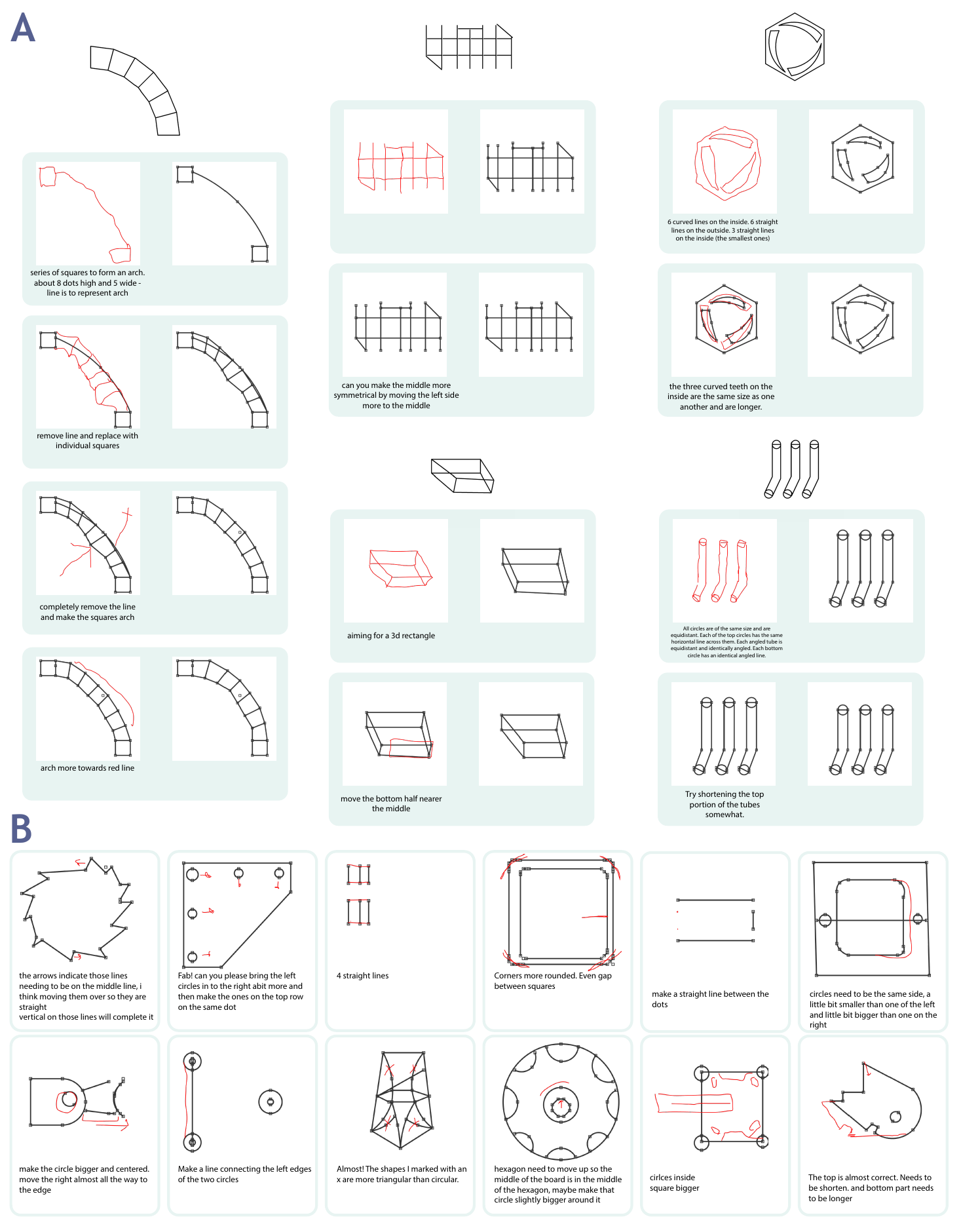} 
  \caption{\textbf{A} Example rollouts from the dataset. Target CADs (top-center) were shown to \textit{Designers}, who created instructions (left columns) that \textit{Makers} followed (right columns). Dyads iteratively refined their CADs across a series of rounds (rows). \textbf{B} Examples of multimodal refinement instructions. Language and drawing mutually constrain and inform the others' semantics. Many instructions don't make sense without the accompanying drawings, and vice-versa.}
  \label{fig:gallery}
\end{figure*}

\section{The \mrcad{} Dataset}
\label{sec:dataset}

\begin{figure*}
  \centering
  \includegraphics[width=1\textwidth]{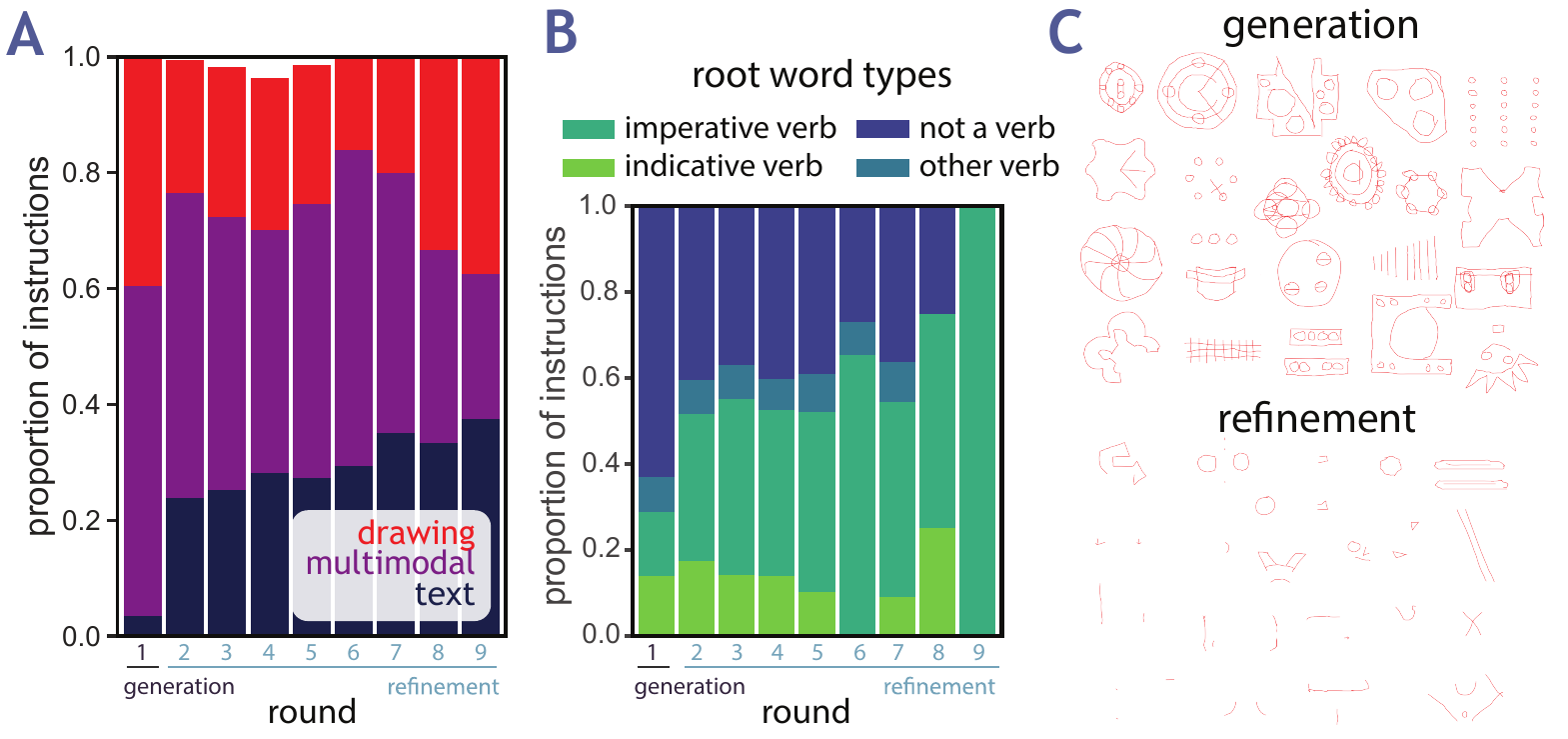}
  \vspace{-2em}
  \caption{\textbf{A} Designers' instructions to \textit{generate} CADs (round 1) involved lots of drawing and little text, whereas instructions to \textit{refine} CADs (rounds 2+) used a balance of modalities. \textbf{B} The proportions of the types of root words in the dependency parse tree of instruction text. More verbs are used over rounds, and these verbs become more imperative. \textbf{C} Samples of 20 generation drawings and 20 refinement drawings highlights the rich detail in generation instructions, and more targeted modifications in refinement.}
  \label{fig:communication_stats}
  \vspace{-1em}
\end{figure*}

We present the main statistics of our dataset, and a set of basic analyses that reveal the striking effects of multi-modality and refinement on communication. We also include a gallery of multi-modal refinement rounds, highlighting the intriguing yet unexplored aspects of the dataset for future research.

\subsection{Dataset statistics}

The full \mrcad{} dataset contains $3166$ unique CADs recreated by $1092$ dyads.
This resulted in a total of 6082 rollouts (for a sample, see Fig. \ref{fig:gallery} A) with a total of $15163$ rounds of instructions and corresponding executions.
On average, each rollout has $2.49$ rounds, and there are a total of $6078$ generation rounds (round 1) and $9085$ refinement rounds (rounds 2+).
To achieve coverage over diverse CADs while also capturing variance in human communication \cite{ji2022abstract}, our dataset contains three distinct subsets: a \textbf{coverage set}, containing $2249$ unique CADs each successfully reconstructed by $1-2$ dyads; a \textbf{dense set}, containing 698 unique CADs each successfully reconstructed by at $3-6$ dyads; and a \textbf{very-dense set}, containing $27$ unique CADs successfully reconstructed by at least $30$ dyads (Fig. \ref{fig:splits}).
In the following analyses we combine data from the coverage and dense sets, and leave the very-dense set for future work.




\subsection{Analyzing instructions in \mrcad{}}

\begin{figure*}[t]
  \centering
  \includegraphics[width=0.9\textwidth]{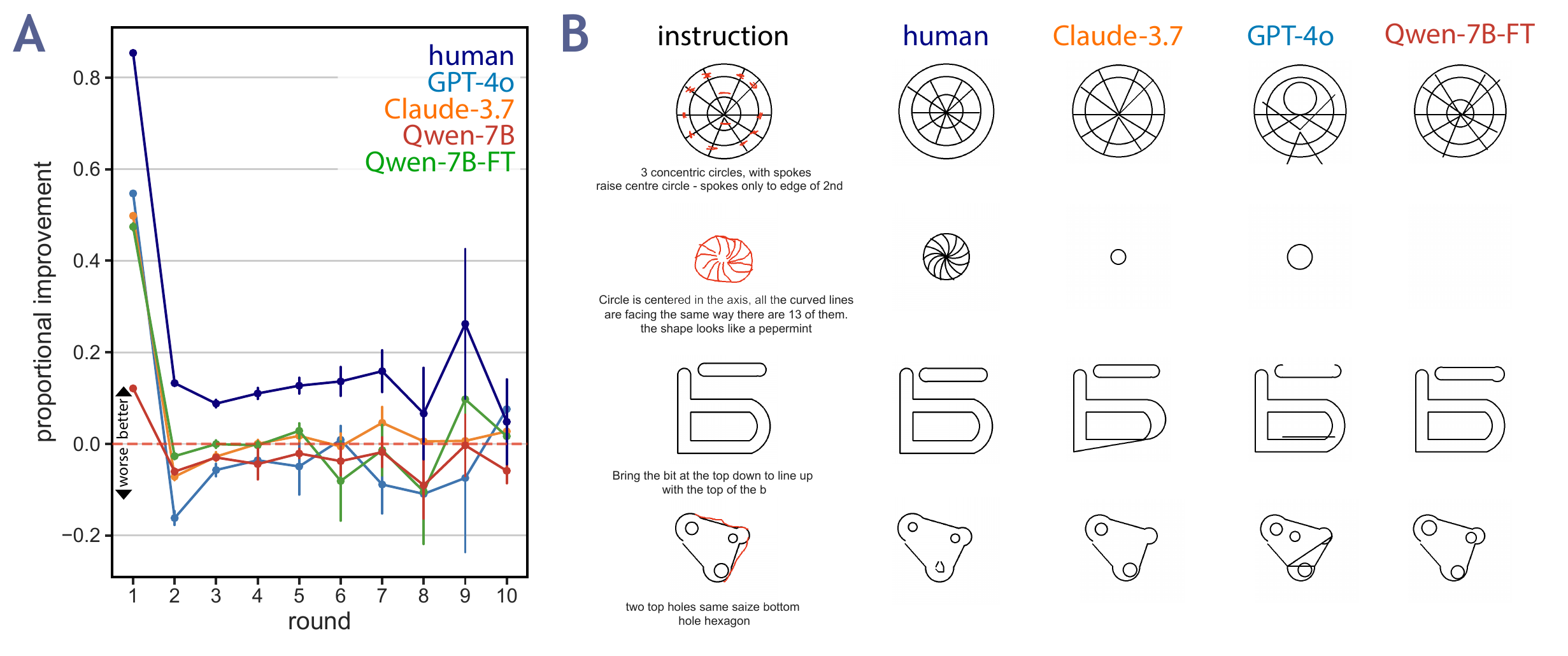} 
  \caption{\textbf{A} Comparison of human and model movement towards target following instructions, normalized by distance at start of round. Only humans make reliably positive changes in responses to refinement instructions. Models made positive steps in generation but largely destructive changes when refining. \textbf{B} Comparison of human and model responses.}
  \label{fig:evals}
\end{figure*}

\paragraph{Refinements increased accuracy of CADs}
A possible concern with implementing a dynamic submission threshold is that dyads may have been able to submit CADs without actually refining their designs, however, we found that, after the initial round, distance to the target continued to decrease ($b=-0.0512$, $t=-16.5$, $p=<0.001$) (Fig. \ref{fig:evals}).
These improvements got smaller as rounds progressed ($b=0.0094$, $t=8.84$, $p=<0.001$). 
Together with the decreasing amount of drawing used across rounds, this suggests that refinements became more fine-grained in later rounds.

\paragraph{Modality use differs in generation and refinement}
Across the 4946 rollouts analyzed, 4125 (83.4\%) contained both text and drawing.
Rollouts containing only drawings made up 14.5\%, and just 2.1\% contained only text, confirming that humans most frequently choose to solve this task using a combination of modalities.
Despite this, many rounds contained only a single modality-- 30.2\% drawing rounds and 16.3\% text rounds.
The distribution of text, drawing, and multimodal instruction was strikingly different in initial generation rounds and subsequent refinement rounds ($\chi^2(2) = 1095, \; p = 0$); almost all generation instructions involved drawings (96.6\%), whereas refinement rounds had a roughly equal split of unimodal text and drawing instructions (Fig. \ref{fig:communication_stats}A).

\paragraph{Refinement drawings are more sparse than generation drawings}
\designer{}s' drawings also changed from generation to refinement rounds.
Most strikingly, \textit{Designers} drew \textit{less} in refinement rounds, as measured by the number of strokes ($b=-5.64$, $t=-72.8$, $p=<0.001$) and amount of digital ``ink'' ($b=20.9$, $t=2.23$, $p=0.026$) used (Fig. \ref{fig:communication_stats}\} B). 
Rendering these images also suggest that the \textit{content} of \textit{Designers}' drawings also changed, from mostly complete renderings of target CADs to smaller modifications of sub-parts of the current CAD.

\paragraph{Refinement text expresses actions and directives} The text of instructions contained an average of $50.2$ characters, or $9.66$ words. To analyze the difference between the text of generation and refinement instructions, we study the root word of the dependency parse tree. We use the English pipeline of the Stanza NLP package \cite{qi2020stanza} to parse the sentences and extract the root words. We tag the root word based on whether it is a verb, and if so whether the mood of the verb is indicative, imperative, or otherwise. We see in Fig.~\ref{fig:communication_stats}B that generation instructions don't have a verb at the head a majority of the time, while refinement instructions do. The proportion of imperatives -- verbs typically used to express directives issued to a listener -- also increases over the course of refinement turns highlighting how refinement instructions have qualitatively different language from generation.

\paragraph{Multimodal refinement messages}
There are a total of 3723 multimodal refinement messages.
Of these, many contained drawings and text that worked together to convey meaning, such that either modality alone would not be sufficient to convey the same intent (Fig. \ref{fig:gallery}B).
This dataset, along with the \textit{very-dense} subset, provide a resource for future work to investigate how language and drawing are used \textit{together} to communicate precisely.


\section{Evaluation}
\label{sec:evaluation}

We evaluate both API access models and open-weights models (which we finetune) on the \mrcad{} dataset. 
We are interested in the following.
\begin{enumerate}[label=\textbf{RQ\arabic*},labelindent=0pt,leftmargin=*,itemsep=0pt,parsep=0pt]
    \item What is the difference in performance between people and models in following the instructions?
    \item What is the effect of generation vs refinement in model performance ?
    \item What is the effect of multi-modal vs single-modal in model performance?
\end{enumerate}

\subsection{Benchmark}
We select CADs from the \textbf{dense set} that have 3 or more successful ($\Delta(D_n, D^*) < 0.2$) rollouts.
This results in $682$ unique CADs with $2324$ rollouts, and a total of $5751$ rounds of instruction-execution pairs.
Then, for these selected rounds, we can evaluate VLM agents by having them assume the role of a human maker $P_\text{maker}$.

\paragraph{Metric}
We evaluate the instruction following abilities by proportional improvement \textbf{PI}: how much the distance to the target $D^*$ shrinks as a consequence of an agent's actions.
\begin{align*}
    \text{PI}(A, R_{:i}, D^*) &= \frac{\Delta(D_i, D^*) - \Delta(A(D_i), D^*)}{\Delta(D_i, D^*)}
\end{align*}
This metric is a proportion of the remaining distance to target at every round.
This accounts for the fact that distances to the target decrease significantly as rounds progress, making late-stage refinements less noticeable without normalization.

\subsection{Evaluating vision-language models}

We build a gym framework for standardized evaluations across models. The gym framework allows for instantiating agents that interface with a standardized representation of CADs, and allows for handing the mrCAD dataset, evaluating models, and simulating interactions between designer and maker agents.

We use the gym framework to evaluate multiple vision-language chat models -- both API access and open weights models -- on the mrCAD benchmark, by having these models assume the role of a human maker.
    \begin{align*}
        P_\text{model-maker}(A ~|~ m, D_i, R_{1:i-1})
    \end{align*}

\paragraph{API access models} 
Since we present the rendered interaction history and drawing instructions to the model, we require models that are accept interleaved image-text inputs. We choose the following models off-the-shelf based on this criterion: GPT-4o \cite{openai2024gpt4ocard}, GPT-4o-mini \cite{openai2025gpt4omini}, Claude-3.7-Sonnet \cite{anthropic2025claude37}, and Qwen2.5-VL-7B-Instruct \cite{bai2025qwen25vltechnicalreport}.
To generate editing actions, we prompt the models to generate editing actions as tool calls. 

\paragraph{Finetuned model}
We also fine-tune a Qwen2.5-VL-7B-Instruct model in the same setting to generate editing actions. We use the 2684 rollouts from the coverage set as a training set. We also include rollouts that don't pass the evaluation threshold. We create a small held-out validation set by choosing 303 successful rollouts from the dense set for 187 CADs that have fewer than 3 successful rollouts for model selection. We train LoRA \cite{hu2022lora} adapters for both vision and language modules of the network.

\begin{table}[]
    \centering
    \begin{tabular}{lrr}
        \toprule
         & \textbf{generation} & \textbf{refinement} \\
        \midrule
        Human & $ 0.854 $ & $ 0.119 $ \\
        \midrule
        GPT-4o & $ 0.547 $ & $ -0.119 $ \\
        $-\text{drawing}$ & $ 0.508 $ & $ -0.159 $ \\
        $-\text{text}$ & $ 0.540 $ & $ -0.124 $ \\
        \midrule
        Qwen-7B FT & $ 0.474 $ & $ -0.017 $ \\
        $-\text{drawing}$ & $ 0.437 $ & $ -0.038 $ \\
        $-\text{text}$ & $ 0.470 $ & $ -0.035 $ \\
        \midrule
        GPT-4o-mini & $ 0.308 $ & $ -0.129 $ \\
        Claude-3.7 & $ 0.498 $ & $ -0.051 $ \\
        Qwen-7B & $ 0.121 $  & $ -0.050 $ \\
        \bottomrule
        \end{tabular}
        \caption{Proportional improvement results for vision-language chat models, including results on ablated instructions. Claude results are from Claude-3.7-Sonnet, and Qwen results are from Qwen2.5-VL-7B-Instruct. Qwen-7B FT is a Qwen2.5-VL-7B-Instruct model fine-tuned on mrCAD data. 
        All models performed worse than humans at following instructions, especially refinement instructions, where the models' outputs made designs worse.
        }
    \label{tab:ft_results}
\end{table}

\paragraph{Ablation on modality}
We ablate one modality (if present) from all the instructions and evaluate them to create $-\text{drawing}$ and $-\text{text}$ evaluation sets. We also perform these ablations on the training data and fine-tune separate Qwen2.5-VL-7B-Instruct models to obtain $\text{Qwen-7B FT}_{-\text{drawing}}$ and $\text{Qwen-7B FT}_{-\text{text}}$ models to 
evaluate on the $-\text{drawing}$ and $-\text{text}$ evaluation sets respectively. We use the same LoRA fine-tuning setup for these experiments.

\subsection{Performance of vision-language models}
For the results, we group rounds as either generation (round 1) and refinement (round 2+). 
Note that this grouping is imprecise and can be improved in future works, as participants do instruct each other to generate in later rounds.

\textbf{RQ1, RQ2} We found (Table~\ref{tab:ft_results}) that models are generally able to decrease the distance to the target during \textit{generation}, with more capable models (GPT-4o, Claude-3.7-Sonnet) and fine-tuned models doing better. However, we find that models perform quite poorly in \textit{refinement} turns. We see that models often make changes that actually \emph{increase} the distance to the target -- in opposition to the intention of the designer's instruction. 
We also see that while an approach like supervised fine-tuning leads to significant improvements in generation turns, the gains don't transfer to refinement turns. This suggests that more sophisticated training approaches might be needed to tackle refinement.

\textbf{RQ3} We also see that models are sensitive to both instruction modalities (Table~\ref{tab:ft_results}), with performance descreasing when a multimodal instruction is ablated to have only a single modality. This effect is more pronounced in generation rounds.




\section{Related Work}

\paragraph{Multimodality}
The instructions in \mrcad{} are multi-modal, consisting of both texts and drawings. 
This is in contrast to other works such as \citet{pejsa2016room2room} and \citet{ku-etal-2020-room}, where the context of the agent is multi-modal, yet the instructions themselves only contain language.
Our drawings efficiently convey spatial information and reference, and even short utterances -- ``delete this'', ``move to here'' -- can hugely constrain the meaning of drawings.
Our work highlights the capacity of drawing to act, not only as a depictive medium, but as a versatile tool for \textit{communication} \cite{goodman1976languages, huey2021semantic, fan2023drawing}, and provides a resource for studying how language is used in conjunction with another medium of communication in a grounded way \cite{lachmy2022draw}.

\paragraph{Generating Designs} 
Prior work has explored AI agents that assist with computer aided design tasks, including those that leverage various kinds of input modality \cite{sanghi2022clip, chen2024generic}, including drawings \cite{seff2021vitruvion}. 
Beyond CAD, other work has explored how agents can support various kinds of designs given multimodal inputs, including generating HTML and CSS code \cite{si2024design2code} and slide-shows \cite{ge2025autopresent}. 

\paragraph{Refinement}
The interactions between people in \mrcad{} takes multiple turns, as a dyad collaboratively refine a current CAD toward a specific target. 
\citet{lachmy2022draw} study complex instructions, but these are presented all at once by the speaker and verified with interpretation by another agent. \mrcad{} prioritizes refinement, where people must elaborate on and repair previous instructions, in this regard, our work is similar to \citet{kim-etal-2019-codraw}. Concurrent work by \citet{zhou2025sweetrltrainingmultiturnllm} presents a reinforcement learning algorithm to train LLM agents to engage in multi-turn conversations with a simulated user for collaborative design.


\section{Discussion}

We present \mrcad{}, a large-scale dataset of multimodal instructions for generating and refining CADs, and corresponding executions in a CAD environment.
Analysis of these data revealed a difference between instructions to \textit{generate} and instructions to \textit{refine}, CADs, notably a trend towards the complementary use of drawing and text when refining designs.
This distinction is mirrored by a performance gap in model instruction-following for generation and refinement. 

We posit that this is partly due to the lack of multimodal refinement data on the internet, from which these models are trained.
While the internet is a rich source of textual and image data, drawing data is much sparser -- particularly the ephemeral drawings that accompany language.
Furthermore, people typically upload only finished artifacts to the internet, often with accompanying contextual text data, making these datasets well-suited for generation tasks.
On the other hand, the process of editing artifacts based on further instructions is rarely uploaded, which is why these models struggle with interactive refinements.
As a multimodal refinement dataset, \mrcad{} serves as a valuable resource for investigating these phenomena.



\bibliography{tacl2021}
\bibliographystyle{acl_natbib}

\onecolumn
\appendix
\section{Dataset collection}

\subsection{Dataset collection interface}
\begin{figure*}
  \centering
  \includegraphics[width=1\textwidth]{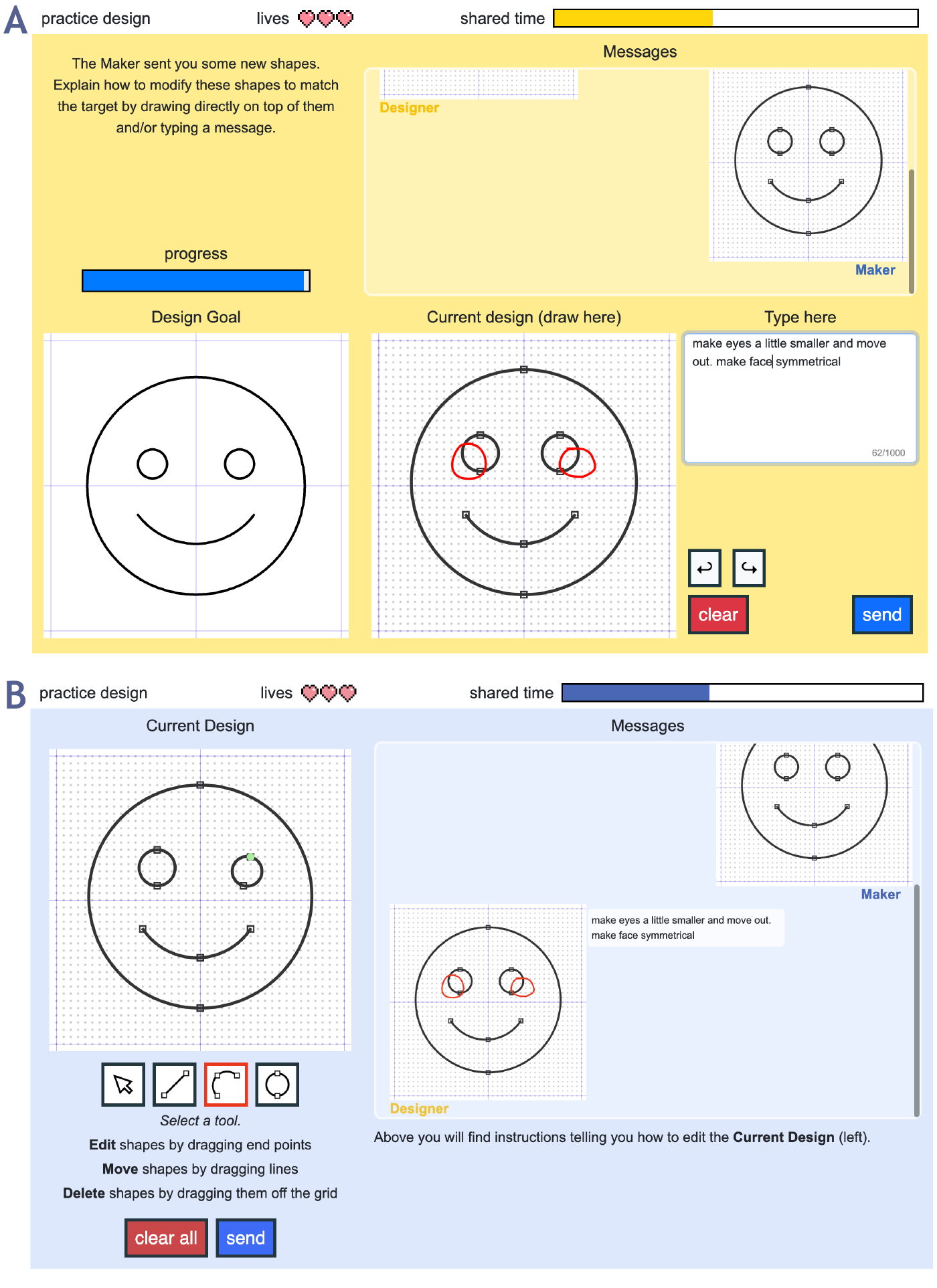}
  \vspace{-2em}
  \caption{\textbf{A} \Designer{} display: target design is shown on the lower leftand instructions are created on the lower right, by drawing on the current CAD and by typing in the text box. History of prior interaction is shown in top right. \textbf{B} \Maker{} display: on the \Maker{}'s turn, the current CAD can be edited in CAD interface on the left, by dragging elements and control points, and by placing new elements by selecting a tool and clicking on grid squares to place control points. Instructions are shown in chat window on right. Both \Designer{} and \Maker{} can see shared time and number of lives remaining. This target was shown to all participants as a practice trial.}
  \label{fig:interface}
  \vspace{-1em}
\end{figure*}

Figure~\ref{fig:interface} shows screenshots of the user interface used for data collection for both players.

\subsection{Dataset collection parameters}
While the key features of our data collection paradigm were kept consistent between our controlled studies (Section \ref{sec:effects-of-multimodality}) and full dataset collection (Section \ref{sec:dataset}), we implemented several changes between versions to increase efficiency of data collection.

\begin{table}[h]
    \centering
    \begin{tabular}{lrr}
        \toprule
         & \textbf{studies} & \textbf{dataset}\\
        \midrule
        Designer turn time & 60s (180s) & unlim. \\
        Maker turn time & 120s (360s) & unlim. \\
        Trial time limit & unlim. & up to 540s \\
        \midrule
        Base payment & \$10 & \$9.10 \\
        Trial bonus & 0 & \$1.00 \\
        Lives & N/A & 3 \\
        \midrule
        Trials per dyad & 6 & 1-13 \\
        Rounds per trial & 3 & 1-10 \\
        \midrule
        Character limit & 200 (600s) & unlim. \\
        Drawing limit & unlim. & unlim. \\
        \bottomrule
        \end{tabular}
        \caption{
            Parameters for ablation studies (generation-only in parentheses) and primary dataset collection. In experiments, we manipulated access to modalities and refinement rounds (across participants). For dataset collection, participants had up to 540 seconds per trial, but time elapsed at double speed for the \Designer{} to discourage precisely drawn copies of the target.
        }
    \label{tab:ft_results}
\end{table}

\section{Analysis}
\subsection{Statistical methods}
Our primary statistical method was the linear mixed effects model.
For modality experiments, we fit models with fixed effects for round and condition, as well as their interaction, and random intercepts for dyad. 
We then compared this full model to a series of simpler, nested models with predictors removed.
We report results from the most complex model for which AIC substantially dropped, compared to the subsequent simpler model.
For the effect of refinement, we we ran paired t-tests between the \textbf{\textit{multimodal + refinement}} condition and the \textbf{\textit{multimodal + generation only}} condition.

\subsection{Distance metric}

The distance between two designs $D, E$, are calculated as follows.

\begin{lstlisting}[style=pythonstyle]
def dist_chamfer_symmetric (D, E):
  return 0.5 * (dist_asymmetric(D, E) + dist_asymmetric(E, D))

def dist_asymmetric(D, E):
  pts_in_D = [sample_points(curve) for curve in D].flatten()
  pts_dists_to_E = [dist_pt_to_design(pt, E) for pt in pts_in_D]

def dist_pt_to_design(pt, E):
  return min([dist_pt_to_curve(pt, curve) for curve in E])
  
\end{lstlisting}

The distance between a point and a curve (line 10) is omitted for brevity, for details please consult the code base.
The default value for the min operator (line 10) is $\frac{1}{4}$, expressing the idea that beyond half a quadrant away, two points are not likely to be unrelated. 

In the analysis, we found a small error in our javascript implementation of the accuracy function that led to a proportion of lower performing trials receiving greater distances than they should have.
As accuracy for inclusion of trials was recalculated post-hoc, entries in our dataset were unaffected.
However, some participants may have performed extra rounds of modification to meet the target threshold. 

\section{Vision-language model experiments}
\subsection{Prompt}
\begin{figure*}
    \small
    \begin{tcolorbox}[colback=red!10, colframe=red!30]
    \texttt{You are an expert CAD software user playing a game called mrCAD. In this game, there is a designer and a maker. The two players work together to iteratively create a design over a sequence of turns. You will play the role of the maker in this game, and the user will play the role of the designer. In each turn the designer provides an instruction about how to modify the design on the canvas. The instruction may include language instructions, drawings on the canvas, or both. The drawings appear as red strokes on the canvas. The design appears in black strokes on the canvas. Your goal is to follow the designer's instructions. You have to take actions to edit the current state of the design. Each action is taken by calling a tool that performs the action. Each control point is a pair of floating point numbers between -20 and 20 that represent the coordinates of the point on the canvas.}
    \end{tcolorbox}
    \begin{tcolorbox}[colback=orange!10, colframe=orange!30]
    \texttt{New game:}
    \end{tcolorbox}
    \begin{tcolorbox}[colback=orange!10, colframe=orange!30]
    \texttt{Round 1. }
    \end{tcolorbox}
    \begin{tcolorbox}[colback=yellow!10, colframe=yellow!30]
    \includegraphics[width=0.15\textwidth]{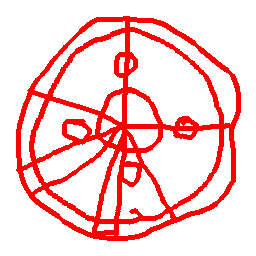}
    \end{tcolorbox}
    \begin{tcolorbox}[colback=yellow!10, colframe=yellow!30]
    \texttt{big round circle with a smaller circle inside it sort of resembling a clock }
    \end{tcolorbox}
    \begin{tcolorbox}[colback=orange!10, colframe=orange!30]
    \texttt{Edit the design based on the designer's instructions using the provided tools. Make sure to follow the instructions carefully.}
    \end{tcolorbox}
    \begin{tcolorbox}[colback=blue!10, colframe=blue!30]
    \texttt{[\{"name": "make\_curve", "arguments": \{"type": "circle", "control\_points": [[0.0, -18.0], [0.0, 18.0]]\}\}, \{"name": "make\_curve", "arguments": \{"type": "circle", "control\_points": [[0.0, -15.0], [0.0, 15.0]]\}\}, \{"name": "move\_point", "arguments": \{"point": [0.0, -15.0], "new\_point": [0.0, -16.0]\}\}, ... ]}
    \end{tcolorbox}
    \begin{tcolorbox}[colback=orange!10, colframe=orange!30]
    \texttt{The resulting design is:}
    \end{tcolorbox}
    \begin{tcolorbox}[colback=gray!10, colframe=gray!30]
    \includegraphics[width=0.15\textwidth]{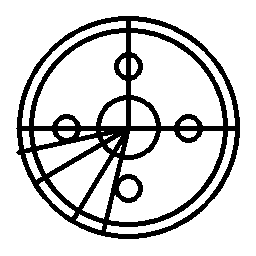}
    \end{tcolorbox}
    \begin{tcolorbox}[colback=gray!10, colframe=gray!30]
    \texttt{\{"curves": [ \{"type":"circle","control\_points":[[0.0,-18.0],[0.0,18.0]]\}, \{"type": "circle", "control\_points":[[0.0,-16.0],[0.0,16.0]]\}, \{"type": "circle", "control\_points": [[0.0,-12.0],[0.0,-8.0]]\}, ...]\}}
    \end{tcolorbox}
    \caption{Structure of the prompt used for API calls as well as SFT models. The prompt includes \colorbox{red!10}{system messages}, \colorbox{orange!10}{templated parts of the instruction presented as user messages}, \colorbox{yellow!10}{\Designer{} instructions}, \colorbox{blue!10}{\Maker{} responses as assistant messages}, and \colorbox{gray!10}{environment feedback presented as user messages}. For Qwen models, a description of the tools is presented as part of the system message (not shown here). For other models, the tool descriptions are integrated by the server.}
    \label{fig:prompt}
\end{figure*}

Figure~\ref{fig:prompt} shows the structure of the prompts used in experiments.

\subsection{Hyperparameters}

\begin{table}[!h]
    \centering
    \begin{tabular}{lcc}
    \toprule
    & \textbf{Qwen-7B(-Instruct and -FT)} & \textbf{others} \\
    \midrule
    temperature & 0.7 & 1 \\
    top-$p$ & 0.95 & 1 \\
    \bottomrule
    \end{tabular}
    \caption{Prompting hyperparameters}
    \label{tab:prompting_hp}
\end{table}

\begin{table}[!h]
    \centering
    \begin{tabular}{lc}
    \toprule
    & \textbf{value} \\
    \midrule
    LoRA parameters & all linear layers \\
    LoRA rank & 128 \\
    LoRA $\alpha$ & 32 \\
    learning rate & $10^{-4}$ \\
    warmup steps & 10 \\
    batch size & 1 / GPU $\times$ 4 GPUs = 4 \\
    gradient accumulation steps & 4 \\
    \bottomrule
    \end{tabular}
    \caption{Fine-tuning hyperparameters for all runs}
    \label{tab:ft_hp}
\end{table}

\end{document}